\documentclass[letterpaper, 10 pt, conference]{ieeeconf}  

\IEEEoverridecommandlockouts                              

\usepackage{graphics} 
\usepackage{epsfig} 
\usepackage{mathptmx} 
\usepackage{times} 
\usepackage{amsmath} 
\usepackage{amssymb}  
\usepackage{soul,color} 
\usepackage{algorithm2e}
\RestyleAlgo{ruled}

\usepackage[english]{babel}
\usepackage{blindtext,tikz}
\usetikzlibrary{calc}

\title{\LARGE \bf
Sample-efficient Model Predictive Control Design of Soft Robotics by Bayesian Optimization 
}

\author{Anuj Pal, Tianyi He$^{*}$, Wenpeng Wei
	\thanks{}
	\thanks{Anuj Pal and Wenpeng Wei are with the Department of Mechanical Engineering, Michigan State University, East Lansing, Michigan, 48824, USA. Emails: {\tt\small palanuj@msu.edu, weiwenpe@msu.edu}. Tianyi He is with the Department of Mechanical and Aerospace Engineering, Utah State University, Logan, Utah, 84342, USA.
		Email: {\tt\small tianyi.he@usu.edu}. }%
	\thanks{* Corresponding author. This work has been submitted to ACC 2023.}
}

\begin{document}
\maketitle

  

\begin{abstract}
This paper presents a sample-efficient data-driven method to design model predictive control (MPC) for cable-actuated soft robotics using Bayesian optimization. Instead of modeling the complex dynamics of the soft robots, the proposed approach uses Bayesian optimization to search the \textit{best-guessed} low-dimensional prediction model and its associated controller to minimize the objective function of closed-loop responses. The prediction model is updated by Bayesian optimization from the closed-loop input-output data in each iteration. A linear MPC is then designed based on the updated prediction model, and evaluated based on the closed-loop responses. Different from directly searching controller parameters, the closed-loop system stability, and inputs/outputs constraints can be easily handled in the MPC design. After a few iterations, a convergent solution of a (sub-)optimal controller can be obtained, which minimizes the user-defined closed-loop performance index. The proposed method is simulated and validated by a high-fidelity simulation of a cable-actuated soft robot. The simulation results demonstrate that the proposed approach can achieve desired tracking controller for the soft robot without a prior-known model. 
\end{abstract}

\section{INTRODUCTION}
Soft robots are made of continuously deformable materials or structures to mimic the biological continuum motions. The distributed softness or continuum brings unique compliance and flexibility compared to the conventional rigid robots. Therefore, the soft robots are showing advantages in the applications of medical devices~\cite{runciman2019soft}, human-robot interactions~\cite{polygerinos2017soft}. However, the softness leads to complex dynamics, which challenges the development of appropriate control algorithms to exploit its advantages. 

The soft robot is ideally an infinite-dimensional system. The continuum structure theory~\cite{trivedi2008geometrically} can accurately describe the dynamics using PDE. However, these models are infinite-dimensional, so they are hard to be used for controller design. In the model-based control approach, the key challenge is to develop a low-dimensional model but accurate enough to achieve good control performance. Many modeling methods of finite-dimensional approximations are reported, including Piecewise Constant Strain (PCC) approach \cite{della2020improved}, Variable Strain (VS) approach \cite{boyer2020dynamics}, Finite Element Model (FEM) \cite{grazioso2019geometrically} and its following model-reduction methods \cite{sadati2021reduced} .

With the appropriate models, different control techniques can be applied. Model predictive control can effectively address the constraints of the states and inputs. This technique has been used in a pneumatic humanoid robot \cite{7551190,8954784}. To account for the model uncertainty, robust $H_{\infty}$ control is used to control distributed actuators in a segmented soft robotic arm \cite{doroudchi2018decentralized}. A detailed review of the model-based control of soft robotics can be found in \cite{della2021model}. It is widely recognized that the performances of model-based methods heavily rely on the accuracy of the model. Therefore, the model-based control approach has another obvious challenge in the trade-off between accuracy/performance and adaptability/complexity. 

Recent advances in the data-driven (or learning-based) control method provide an alternative approach to the model-based control approach. Traditional machine learning (ML) or deep learning (DL) methods can be applied to approximate the complex model without a prior understanding of the system dynamics. The input-output data will be used to establish a black-box model. However, the black-box model is hard to be implemented in the control design. The closed-loop system stability is still an unsolved issue. Therefore, its applications in real-time control are limited. A review paper on the machine learning of soft robots provides more details \cite{kim2021review}. 

A promising middle-point method is to integrate data-driven and model-based control to tackle the control of complex soft robots. This emerging method is considered a powerful candidate to address the complexity and nonlinearity. Some works of such methods are reported in the literature. An iterative learning model predictive control can improve the model accuracy by gradually updating the model parameters using the data from repetitive processes \cite{8793871}. Koopman operator theory offers a data-driven method to generate a linear model in the lifted high-dimensional space to approximate the nonlinear model. Linear control methods, including linear MPC and LQG, can then be easily implemented  \cite{bruder2019nonlinear,bruder2020data}. However, both the ML/DL and Koopman operator approaches need a large number of experiments, collecting data to learn a good model. This drawback leads to expensive costs in the learning process and hinders the applications of soft robots.  

Bayesian optimization is a promising approach that has been proven to reduce the computational burden of identifying the optimal parameters for any complex system. The approach involves the use of a data-driven model and an actual system to iteratively improve the system knowledge as per the desired performance function. The approach has been validated in easing the computational burden for various problems ranging from parameter calibration and control design. References \cite{pal2019engine, zhu2020engine, 9329078, pal2021multi, 9147983, gutjahr2017advanced, tang2021stochastic} implemented the Bayesian optimization framework in automotive domain for performing the engine calibration. Apart from automotive applications, the Bayesian optimization approach has also been successfully implemented in other applications such as analog/rf circuit design \cite{lyu2018multi}, groundwater reactive transport model \cite{zhou2018adaptive}, actuator modeling \cite{liu2017efficient}, and designing natural-gas liquefaction plant \cite{ali2018surrogate}. All these works have shown the capability of data-driven approaches for modeling a complex system with a relatively simple model for computation.

 
In this paper, we present a sample-efficient data-driven method to design the MPC of a cable-actuated soft robot by Bayesian optimization. Due to the nonlinearity and complexity of the soft robot, the nonlinear mapping from the model parameters and associated control parameters to ultimate system performance is hard to model. We treat the nonlinear mapping as a Gaussian process and use Bayesian optimization to find the best prediction model that matches with online input-output data. The \textit{best-guessed} prediction model is then used to design a linear MPC. The closed-loop data is collected to evaluate the control performance and update the prediction model by Bayesian optimization. An optimal solution of the \textit{best-guessed} model and MPC will be obtained after iterating a few experiments.

To the best knowledge of the authors, this is the first time that MPC for the soft robot has been designed using the Bayesian optimization technique. The main contributions of this work are three-fold: 1) Proposing a Bayesian optimization framework for the soft robot in identifying the optimal reduced-order system dynamics approximation 2) Integrating linear MPC with Bayesian optimization to achieve accurate tracking control of soft robots 3) Demonstrating the tracking control performance and computational burdens of the proposed method. 

The rest of this paper is organized as follows. Section~\ref{problem_formulation} formulates the problem and shows the overview of the control scheme. Section~\ref{Bayesian_opt} introduces Bayesian optimization and the algorithm for control design.   Section~\ref{simulation} then presents the simulation results in a high-fidelity environment. At last, conclusions are made, and future work is discussed.

\section{Problem formulation}\label{problem_formulation}
Consider a soft robot actuated by cables, which is a multi-input-multi-output (MIMO) system. Its nonlinear dynamic is described by the discrete-time nonlinear system \eqref{nonlinear_model}
\begin{equation}\label{nonlinear_model}
 	\begin{array}{c}
	{x}_{t+1} = f(x_{t},u_{t}) \\
y_{t} = h(x_{t}, u_{t})
	\end{array}
\end{equation}
where $x_{t}, u_{t}, y_{t}$ denote the state, control inputs and outputs at time index $t$. The nonlinear function $f(\cdot)$ and $h(\cdot)$ are the nonlinear dynamic model of the soft robot that is assumed to be unknown. 

The soft robot is actuated by three cables embedded in the body, and the end point is installed by a laser. The end-point will point at the $x-y$ plane, and the tracking controller is expected to track the reference trajectory on the $x-y$ plane. 

 \begin{figure}[ht]
 	\centering
 	\includegraphics[clip, trim=6cm 3cm 6cm 3cm, width = 0.75\linewidth]{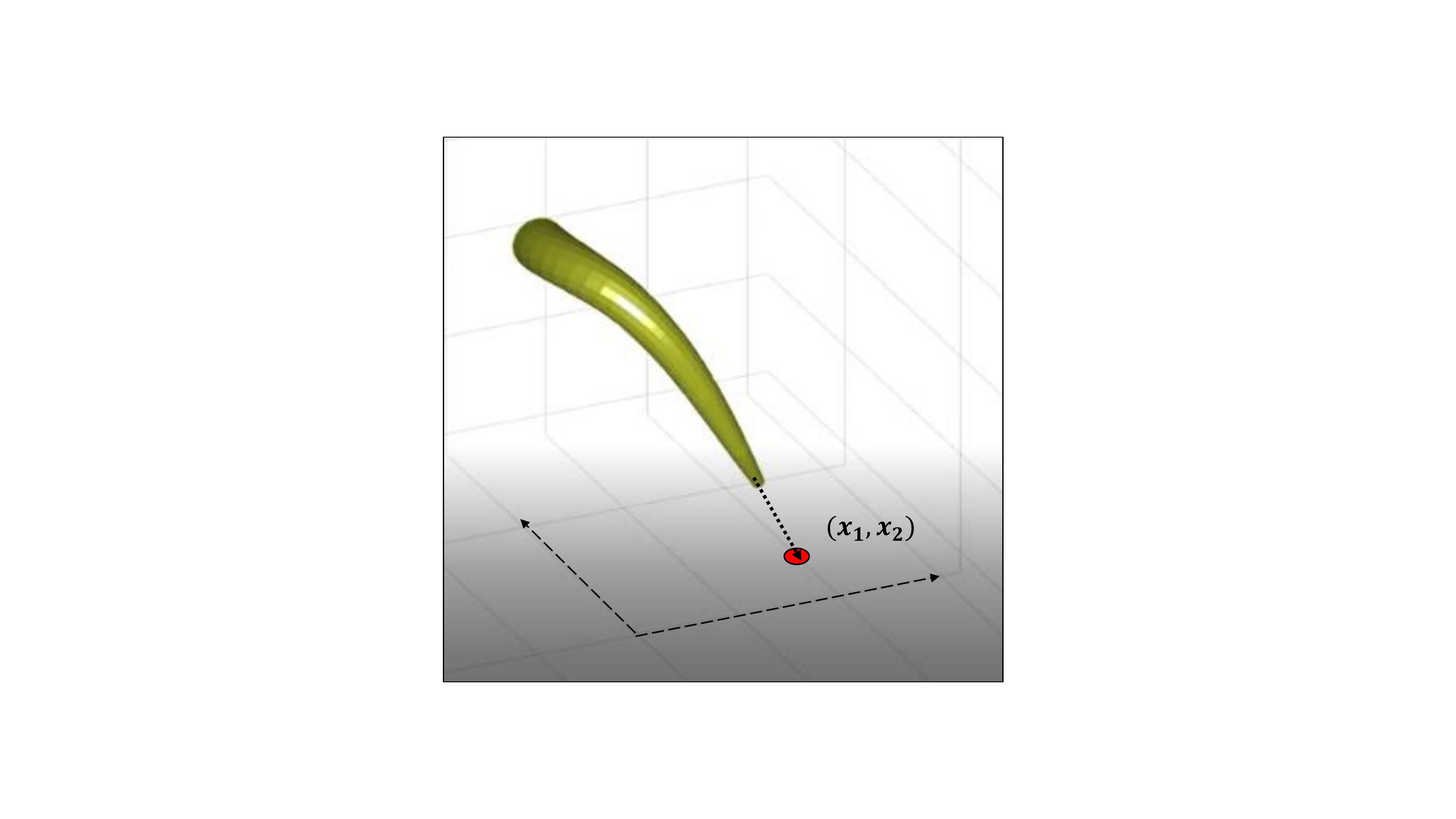}
 	\caption{Cable-actuated soft robot and positioning of its end-point on 2D plane.}
 	\label{fig:control_scheme}
 \end{figure}

The pre-defined task is to track a given reference trajectory $r_{t}$ in the time horizon $(t = 1, 2, \dots, N)$, written as $[1, N]$. The control law \eqref{control law}
\begin{equation}\label{control law}
	u_{t} = g(y_{t}, r_{t})
\end{equation}
is expected to enforce $y_{t}$ tracks reference $r_{t}$. The objective is to find a control law that minimizes the weighted tracking errors and control inputs within the entire time horizon, as indicated in the problem formulation in \eqref{cost_function}. $Q, R$ are weighting matrices for tracking errors and inputs, and $Q \geq 0,~ R > 0$.
\begin{subequations}\label{cost_function}
	\begin{align}
	\min _{\{u_{t}\}_{t=1}^{N}} J & = \min _{\{u_{t}\}_{t=1}^{N}} \sum_{t=1}^{N} (y_{t}-r_{t})^{T}Q (y_{t}-r_{t}) +u_{t}^{T} R u_{t} \\
		\text{subject to: }
		& \text{unknown}: {x}_{t+1} = f(x_{t},u_{t}) \\
		& \text{unknown}:    y_{t} = h(x_{t}, u_{t})\\ 
		& u_{min}  \leq u_{t} \leq u_{min} \\ 
	    & y_{min} \leq y_{t} \leq y_{min} 
	\end{align}
\end{subequations}

The optimal solution of the control law is denoted as $u^{*}_{t} = g^{*}(y_{t}, r_{t})$. The unknown dynamics $f(\cdot)$ and $h(\cdot)$ make the mapping from control law $g^{*}$ impossible to be evaluated by the unknown models. In the traditional model-based approach or Koopman operator approach, a tremendous amount of data needs to be collected by conducting sufficiently many experiments at operating points covering the whole range of interest. Therefore, solving the optimal control law is very expensive for the traditional methods that need to establish the mapping from control inputs to the ultimate system performance. However, a given control law can be evaluated by conducting experiments/simulations, collecting the input-output data, and analyzed in the cost function of \eqref{cost_function}. In other words, the control law can be sampled and improved through the online input-output data. The Bayesian optimization is a suitable tool to achieve this control objective without precise dynamic models. 

Instead of using Bayesian optimization to directly auto-tune the controller~\cite{neumann2019data}, Bayesian optimization is used in this paper to improve the approximated prediction model based on the best '\textit{guess}' in each iteration. The well-studied MPC will be used to design the controller, and the performance index will be evaluated on the closed-loop system with MPC. 

  \begin{figure}[ht]
 	\centering
 	\includegraphics[clip, trim=3.5cm 3.5cm 4.1cm 3cm, width = \linewidth]{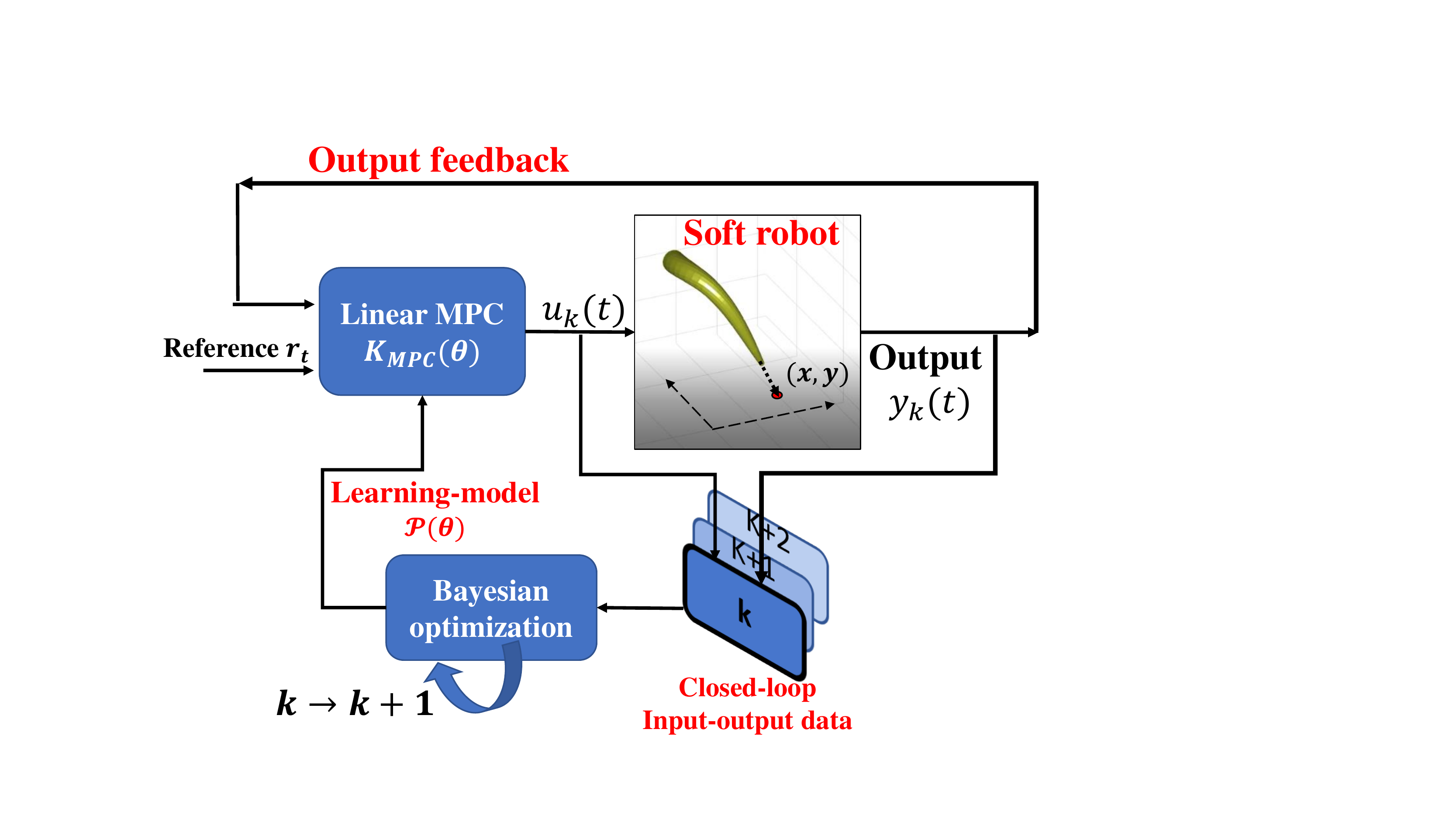}
 	\caption{Overview of tracking control scheme of soft robotics by Bayesian optimization and MPC. $t$ - time index, $k$ - iteration index.}
 	\label{fig:control_scheme}
 \end{figure}
 
 The overview of the proposed control scheme is shown in Fig. \ref{fig:control_scheme}.  The prediction model $\mathcal{P}$ in the state-space representation to approximate the input-output relationship at a local operating point is denoted as 
 \begin{equation}\label{prediction_model}
 		\begin{aligned}
 		    {z}_{t+1} & = A_{p}z_{t} + B_{p}u_{t} \\
 			y_{t} & = C_{p}z_{t} + D_{p} u_{t}
 		\end{aligned}
 \end{equation}
 where $z_{t}$ is the state of the prediction model, whose dimension can be selected. $A_{p}, B_{p}, C_{p}, D_{p}$ are the prediction model matrices, the elements of which are model parameters. For ease of expression, the vector $\theta$ denotes model parameters to be tuned that are stacked from all the matrices elements. The dimension of the prediction model $\mathcal{P}(\theta)$ cannot be selected too high to avoid the curse of dimension. Normally, the dimension of $\theta$ is chosen as $10-20$.  
 
At each iteration, the MPC uses the updated prediction model to design the control inputs in the prediction horizon $T_{p}$. Note that the prediction horizon is usually smaller than the entire time horizon $[1, N]$ of the tracking control task. At each time step, the control input is obtained from the MPC design by solving the following optimization, and the controller is denoted as $K_{MPC}(\theta)$. The control inputs at the entire time horizon can be obtained by repeating solving \eqref{MPC_prediction} until the task is complete. 
 \begin{subequations}\label{MPC_prediction}
 	\begin{equation}
 		\min _{\{u_t\}_{t=1}^{T_{p}}} \sum_{t=1}^{T_{p}} (y_{t}-r_{t})^{T}Q (y_{t}-r_{t}) +u_{t}^{T} R u_{t}
 	\end{equation}
 		\text{subject to: } \text{updated prediction model} $\mathcal{P}(\theta)$
 	 \begin{equation}
 	 	\begin{array}{c}
 	 			u_{min}  \leq u_{t} \leq u_{min}  \\ 
 	 			 y_{min} \leq y_{t} \leq y_{min} 
 	 	\end{array}
 	 \end{equation}
 \end{subequations} 
 
 The unknown system dynamic equations \eqref{cost_function} is replaced by the '\textit{guessed}' prediction model \eqref{prediction_model}, and it will be used to design a linear model predictive control. Using the closed-loop input-output data, the performance in the entire time horizon can be evaluated. The Bayesian optimization is then used to tune the prediction model $\mathcal{P}(\theta)$ such that the input-output data can be well matched. 
 
 The advantages of this method are three-fold: 1) The precise model is not needed, and the prediction model will be improved in each iteration based on the evaluations of closed-loop system performance; 2) The controller subject to the constraints can be designed in a unified way of linear MPC. The closed-loop stability and robustness can be guaranteed by the well-established linear MPC design; 3) This method improves the computational efficiency than existing methods because a relatively low-dimensional prediction model is used in the MPC.
 
\section{Bayesian optimizations }\label{Bayesian_opt}
Bayesian optimization (BO) denotes a class of algorithms for black-box global optimization problems in which data collection is expensive [9], and thus, only a few evaluations are possible. To deal with scarce data, BO (i) assumes a probabilistic prior about the objective function and (ii) chooses wisely the next combination of parameters to try on the system according to a pre-established acquisition function.

Bayesian optimization models the objective function as a GP model and utilizes the model estimations to select the next function evaluation until it finds the global minimum or hits the maximum allowable iteration number. The BO algorithm aims at approaching the optimal solution $\theta^{*}$, by which the linear MPC law $K_{MPC}(\theta)$ optimizes the control performance on the entire task time horizon. $\Theta$ is the search space of the parameter $\theta$.
\begin{equation}
	\theta^{*}=\arg \min_{\theta \in \Theta} J(\theta)
\end{equation}

The evaluation of objective function $J$ is conducted by the closed-loop experiments/simulation, which samples the parameter $\theta$ in the search domain. Two key steps involved in performing the BO are the model development and then formulating the acquisition function to intelligently search for the optimal solution. A Gaussian process model is developed, which predicts both mean and uncertainty quantification. Both these parameters are then combined in a way to perform efficient exploration and exploitation of the design/search space to identify the optimal solution.

\subsection{Gaussian process}
Gaussian process regression (GPR) is used to formulate the latent function \textit{f} capturing the interaction between system performance (cost function $J$) based on input parameters to be optimized ($\theta$).

The expression for the latent function is given as
\begin{equation}
    f \thicksim GP(m,k)
\end{equation}
where, $f$ is the latent function, capturing the input-output behavior, and $m$ and $k$ are mean and covariance. If the system is corrupted with noise, then the output can be written as:
\begin{equation}
    y = J(\theta)  = f + \epsilon, \quad where~ \epsilon \thicksim N(0,\sigma^2_n) 
\end{equation}

Here, $\epsilon$ is the measurement noise considered as Gaussian with zero mean and variance $\sigma^2_n$. Combining both the terms in the above expression, the output \textit{cost} can be written as:
\begin{equation}
    J(\theta) \thicksim GP(m, k + \sigma^2_n\delta_{pq})
\end{equation}
where $\delta_{pq}$ = 1 iff $p = q$ is the Kronecker's delta. The output is a GP with mean m and covariance $k$ + $\sigma^2_n\delta_{pq}$. Several mean and covariance functions are available to choose to develop the model. For this work, m = 0 is selected, which is a very common mean function used for model development. A Gaussian covariance function is selected, which introduces several \textit{hyperparameters} to the model. Hyperparameters are the unknown parameters that are optimized to get the best possible fit of the function to the available data. The optimal hyperparameters are obtained by maximizing the log marginal likelihood function. 

Let us assume that we initially have $N_{p}$ points to fit the Gaussian Process model. Therefore, the output cost function with measurement noise can be written as:
\begin{equation}
    J(\mathbf{\theta}) \thicksim GP(\textbf{m}, \textbf{K} + \sigma^2_n\textbf{I})
\end{equation}
where, $\mathbf{\theta}$ represent an input matrix, \textbf{m} is the vector with mean values and \textbf{K} is the covariance matrix given as $(k(\textbf{x}_i,\textbf{x}_j)) ~\forall~ i,j \in [1,N_{p}]$.

After fitting the model, it is important to perform predictions at any unknown input point. Let $J(\mathbf{\theta})$ be the known outputs, and $J(\theta')$ is the unknown output that we want to predict at input $\theta'$ using the Gaussian Process model. Since the process is Gaussian, we can write the joint distribution of both known and unknown points as:

\begin{equation}
    \begin{bmatrix}
    J(\mathbf{\theta}) \\
    J(\theta')
    \end{bmatrix} \thicksim N\Big(\begin{bmatrix}
    \textbf{m} \\ 
    m(\theta')
    \end{bmatrix},\begin{bmatrix}
     \textbf{K} + \sigma^2_n\textbf{I} & \textbf{k}(\mathbf{\theta},\theta')\\ 
    \textbf{k}^T(\mathbf{\theta},\theta') & k(\theta',\theta'))
    \end{bmatrix}  \Big)
\end{equation}

Here, $J(\mathbf{\theta})$ is the training set, and $J(\theta')$ is the unknown output. $\textbf{k}(\mathbf{\theta},\theta')$ = $k(\theta_i,\theta') ~\forall~i \in [1,N_{p}]$ represent the covariance calculation between known data set $\mathbf{\theta}$ and unknown test point $\theta'$. Using the above expression, the conditional distribution of $J(\theta')$ given $J$ can easily be obtained. For the sake of brevity, we have eliminated the mathematical discussion for the model development. Readers are encouraged to go through the reference \cite{williams2006gaussian} for more details. After performing mathematical calculations, the model estimates at any unknown point ($\theta'$) can be written as: 


\begin{equation}
\begin{array}{rl}
   J_m(\theta')  =&  m(\theta') + \textbf{k}(\mathbf{\theta},\theta)^T\textbf{K}^{-1}(J(\mathbf{\theta}) ~-~ \textbf{m})\\
   J_{\sigma^2}(\theta') =& k(\theta',\theta') - \textbf{k}(\mathbf{\theta},\theta')^T\textbf{K}^{-1}\textbf{k}(\mathbf{\theta},\theta')
\end{array}
\end{equation}

where, $J_m(\theta')$ and $J_{\sigma^2}(\theta')$ are the posterior mean and variance estimate from the Gaussian process at any unknown point $\theta'$. The variance estimate becomes crucial for intelligently identifying the optimal region. The next section discusses the acquisition function, which combines both mean and variance estimates from the model to direct the Bayesian optimization algorithm toward the optimal region.

\subsection{Acquisition function: Expected improvement}
The expected improvement seeks to obtain the next evaluation point, where the objective function is expected to be improved most over the best cost function value $J^{*}$ collected so far. 
One major advantage of the Gaussian process model is its mean and variance estimation. The mean value suggests the region of the optimal solution, whereas the variance estimate shows the uncertainty in unexplored regions. The acquisition function utilizes both mean and variance estimates to balance the exploration-exploitation of the design space, which is crucial for finding global optima. 

There are various ways to formulate the acquisition function [ref]. For this work, we are using Expected Improvement (EI) \cite{jones1998efficient}. It is derived by formulating an improvement function $I(\textbf{x})$ and then taking its expectation. The improvement function for the current problem is defined as:
\begin{equation}
    I(\theta) = max\{0, J_m(\theta) - J^*\}
\end{equation}
where, $J^* = max_{x \in N}J$ is the best value of the cost function obtained so far, and $J_m(\theta)$ is the cost predicted by the Gaussian model. Taking the expectation of the improvement function, we obtain the expression for EI, given as:
\begin{equation}
\begin{aligned}
   \mathcal{A}_{k}(\theta) = EI(\theta) = (J_m(\theta) - J^*)\Phi\Big(\frac{J^* - J_m(\theta)}{J_{\sigma}(\mathbf{\theta})}\Big)  \\
     + ~ J_{\sigma}(\theta)\phi\Big(\frac{J^* - J_m(\theta)}{J_{\sigma}(\theta)}\Big)
\end{aligned}
\label{EI}
\end{equation}
Here, $\Phi$ and $\phi$ are the standard normal cumulative distribution function and probability distribution function, respectively. The goal here is to maximize the EI function, which depends on the mean value and variance estimate from the model, shown in the first and second part of the equation (\ref{EI}), respectively.




The overall Bayesian optimization algorithm to design MPC for soft robot is summarized as follow.  

\begin{algorithm}
	\caption{Bayesian optimization of MPC of soft robots}\label{alg:cap}
	\KwData{$Q \geq 0, R > 0, u_{min}, u_{max}, y_{min}, y_{max}, N >  T_{p} > 0$, search space $\Theta$, null dataset $\Theta_{set}, J_{set}$.}
	\KwResult{$\theta$, i.e. $A_{p}, B_{p}, C_{p}, D_{p}$}
	\For{$k < k_{max}$}{
	\begin{itemize}
	    \item   Update the mean and variance of the objective function $J(\theta)$ 
     \item Optimize the acquisition functin to determine next evaluation point $\theta_{k+1}$ \\
			$\theta^{*} \leftarrow \arg \min_{\theta \in \Theta} J(\theta)$
   \item Design linear MPC and conduct experiments using $K_{MPC}(\theta_{k+1})$ and measure $\hat{J}(\theta)$ 
   \item  Augment the data set \\
 ${\Theta}_{set,k+1} \leftarrow {\Theta}_{set,k} \cup \theta_{k+1}$ \\
$J_{set,k+1} \leftarrow J_{set,k} \cup J\left(\theta_{k+1}\right)$
	\end{itemize}	
}

$	\theta_{o p t}=\theta_{opt, k_{max}}=\arg \min _{\theta \in \Theta} J(\theta) $;
\end{algorithm}

\section{Simulation Results and discussions}\label{simulation}
To validate the effectiveness of the proposed method in the control of soft robots, several simulations are conducted in a high-fidelity simulation environment - \textit{RoBoSim}. The simulation platform can simulate both the static and dynamics of the continuum and soft robots~\cite{renda2020geometric,boyer2020dynamics}. The simulation of the soft robot is based on the Cosserat-beam theory. The beam shape is characterized by the nonlinear parameterization by strain fields, and the order-reduction is based on a functional basis of strain modes. While remaining geometrically exact, the simulation platform provides us with a minimal set of ordinary differential equations that can be easily implemented for analysis and control design. The accuracy of the model is comparably accurate to the well-validated finite element method in the nonlinear structural statics and dynamics.

\subsection{Simulation setup}
The Bayesian optimization is conducted using the Bayesian Optimization Toolbox \textit{bayesopt} by MATLAB. The soft robot in \textit{SoRoSim} and the Bayesian optimization are co-simulated with the sampling time $50$ ms in simulation time of $20$ seconds. The soft material of the robot body is selected as PDMS. The outputs $x_{1}$ and $x_{2}$ displacement are constrained by $[-0.1, 0.1]$. The control input is constrained by $[-10, 10] N$. The parameters of robot and Bayesian optimization are summerized in the Table~\ref{tab:my_label}.
\begin{table}[h]
    \centering
      \caption{Summary of parameters in the simulation}
    \begin{tabular}{c|c||c|c}
    \hline
       BO parameter  & value & soft robot & values\\
       \hline
       white noise variance $\sigma^{2}$ & $0.01^2$ & density & 1000 kg/m$^{3}$\\ 
       maximum iteration $k_{max}$ & 100 &  Young's modulus  & 1e6 N/m$^{2}$\\ 
       number of seed point & 10 &  Poisson's ratio  & 0.5 \\ 
       GPA active set size & 300 &  length  & 0.6 m\\
       prediction horizon $T_{p}$ & 10 & radius  &  0.1 m\\
       weighting matrix $Q$ & $10^{3}$ $I_{3}$ & actuation cable  & 3\\ 
       weighting matrix $R$ & $ I_{3}$   &  damping & 0.11e5 Pa.s
    \end{tabular}
    \label{tab:my_label}
\end{table}
\subsection{Simulation results}

\subsubsection{Scenario 1: position initialization to origin}
In this simulation scenario, the soft robot is driven to position to the origin point to initialize the system. The tracking target is the constant reference trajectory at origin point $(0, 0)$. 
\begin{figure}[ht]
	\centering
	\includegraphics[clip, trim=0.8cm 6.5cm 1.0cm 7.0cm, width = 0.85\linewidth]{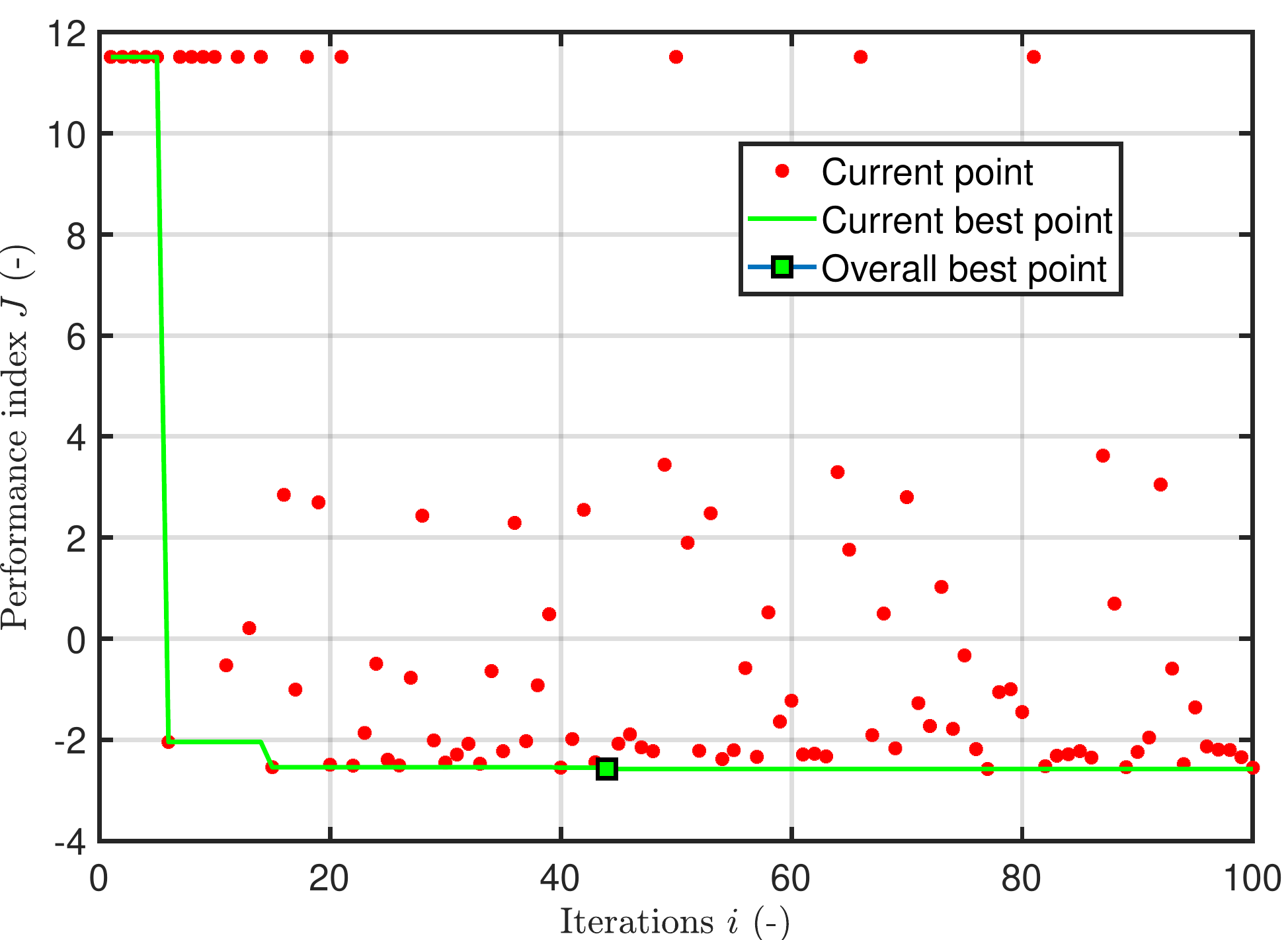}
	\caption{Performance index in iterations of Bayesian optimizaion.}
	\label{fig:Bayesian_opt_iterations}
\end{figure}

The performance index $J$ at the iterations is plotted in the Fig.\ref{fig:Bayesian_opt_iterations}. The best performance point is achieved at the green point of the $42^{th}$ iteration. At the early stage of convergence, a sub-optimal solution can be achieved by around 10 iterations. This indicates that a good controller can be tuned by the Bayesian optimization in a few iterations.

\begin{figure}[ht]
	\centering
	\includegraphics[clip, trim=0.8cm 10.8cm 2.2cm 10.5cm, width = \linewidth]{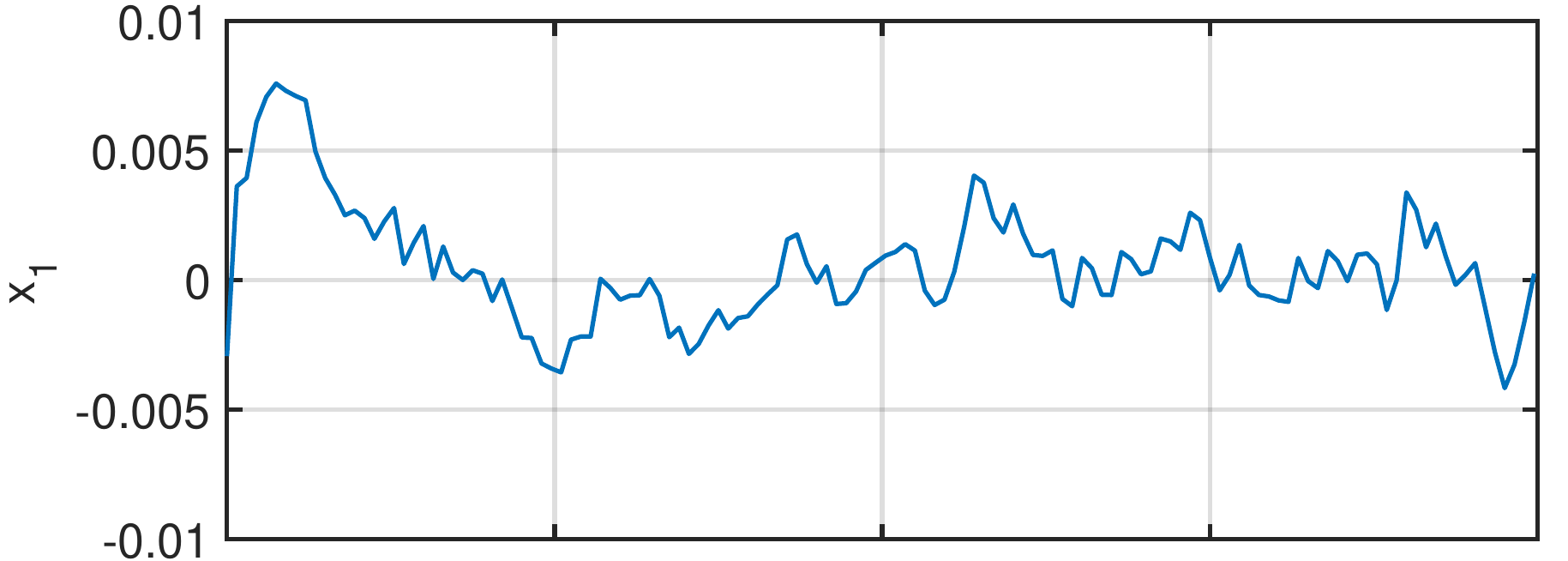}
 \vspace{-2.3 em}
 
    \includegraphics[clip, trim=0.7cm 10.3cm 2.2cm 9.4cm, width = \linewidth]{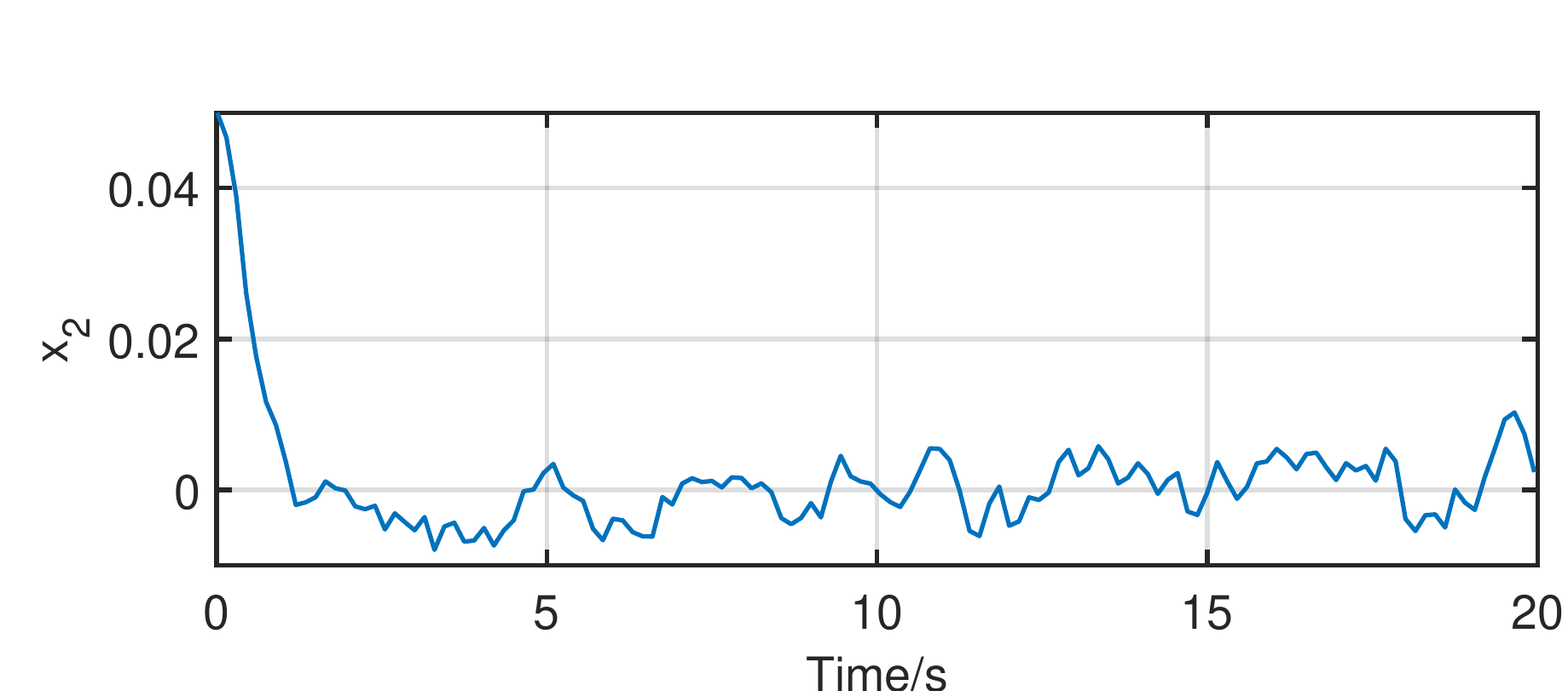}
	\caption{Position tracking of the origin under perturbation.}
	\label{fig:best_evaluation_resp}
\end{figure}

The responses of the best design at $x_{1}$ and $x_{2}$ are shown in Fig.\ref{fig:best_evaluation_resp}. From the initial condition point, the states can converge rapidly to the origin point. Because the measurement is subject to measurement noises, the position responses are fluctuating around 0.  

\subsubsection{Scenario 2: tracking a circular shape}

\begin{figure}[ht]
	\centering
	\includegraphics[clip, trim=0.5cm 6.5cm 1cm 7.2cm, width = 0.9\linewidth]{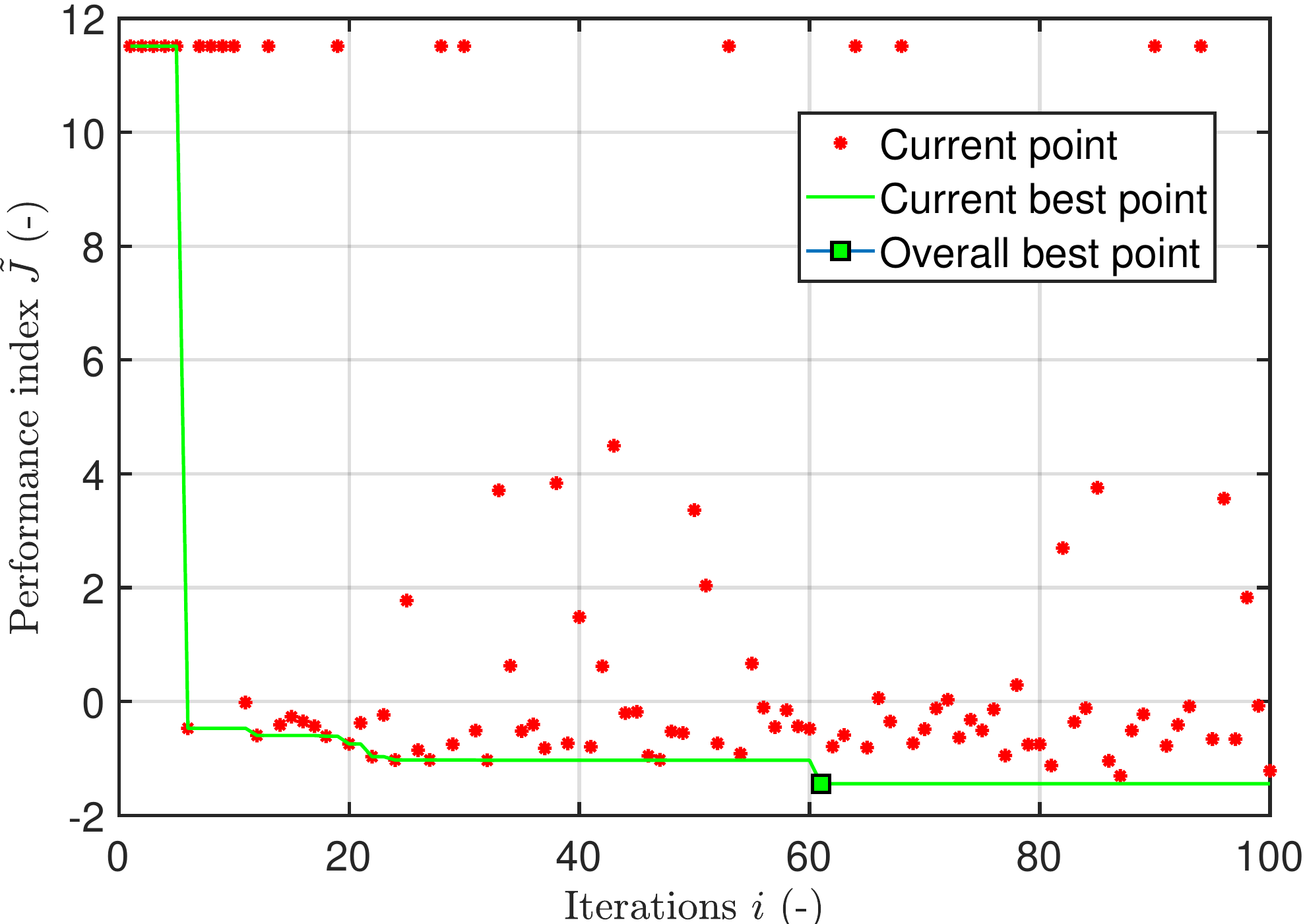}
	\caption{Performance index in iterations of Bayesian optimization.}
	\label{fig:Bayesian_opt_circle}
\end{figure}

The reference trajectory to be tracked is a circle with a radius of $0.05$ m. The equation is expressed as $x_{1}^{2} + x_{2}^2 = 0.05^2$. The convergence of the performance relative to iterations is plotted in Fig. \ref{fig:Bayesian_opt_circle}. Although it takes longer iterations to achieve the optimal, the sub-optimal can be obtained in a few iterations. 

\begin{figure}[ht]
	\centering
	\includegraphics[clip, trim=1.0cm 5.2cm 2cm 6cm, width = 0.85\linewidth]{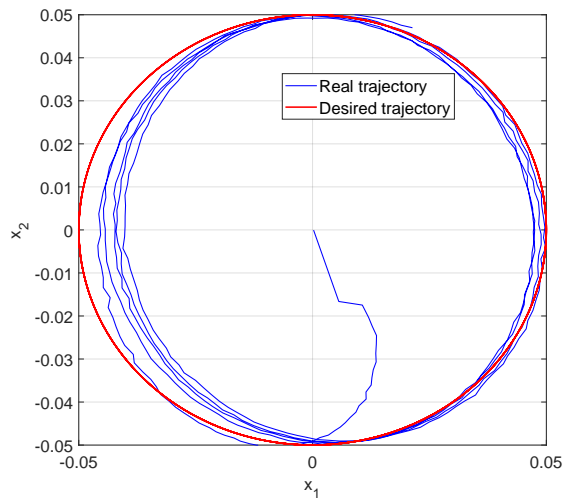}
	\caption{Position trajectory tracking of the desired circular shape trajectory.}
	\label{fig:best_resp_circle}
\end{figure}

The responses of the positions $x_{1}$ and $x_{2}$ are plotted in Fig.\ref{fig:best_resp_circle}. The red curve is the response from the actual soft robot, and the blue curve is the desired circular shape. In the time horizon of the tracking task, the soft robot starts from the origin point and converges to the desired circle. Gradually, the position curve matches the desired trajectory. 

\section{Conclusions and future work}
This paper presents a systematic framework for approximating the complex dynamics of a soft robot using the Bayesian optimization (BO) framework. The study integrates the framework to identify the low-dimensional linear dynamic model, which is then used with a linear MPC controller to achieve the desired tracking performance. For BO, the Gaussian process is used for model development, and expected improvement (EI) is used as an acquisition function to perform exploration and exploitation of the search/design space to obtain optimal solutions. The results demonstrate superior performance using our proposed approach in identifying the approximate system dynamics, which is then used with linear MPC to achieve desired system performance. The future work includes investigating the robustness of the proposed method and experimental validations by real-time implementations.


\bibliographystyle{IEEEtran}
\bibliography{reference}

\end{document}